\documentclass[Journal]{IEEEtran}

\usepackage{amssymb,color,amsmath}
\usepackage{paralist,cite}
\usepackage{graphicx}

\usepackage{epstopdf}

\newcommand{\nix}[1]{}

\begin{document}
\title{Hajj and Umrah Event Recognition Datasets}

\author{
{Hossam Zawbaa${^\dag}$~~~~~~~~~~~~~~~~~Salah A. Aly$^{\dag\ddag}$}\\
\medskip
{$^{\dag}$Center of Research Excellence in Hajj and Umrah, Umm Al-Qura University, Makkah, KSA\\ $^{\ddag}$College of Computer and Information Systems, Umm Al-Qura University, Makkah, KSA\\
Email: salahaly@uqu.edu.sa}}
\maketitle

\begin{abstract}
In this note,  new Hajj and Umrah Event Recognition datasets (HUER) are presented. The demonstrated datasets are based on videos and images taken during 2011-2012 Hajj  and Umrah seasons. HUER is the first collection of datasets covering the six types of Hajj and Umrah ritual events (rotating in Tawaf around Kabaa, performing Sa'y between Safa and Marwa, standing on the mount of Arafat, staying overnight in Muzdalifah, staying two  or three days in Mina, and throwing Jamarat). The HUER datasets also contain video and image databases for nine types of human actions during Hajj and Umrah (walking, drinking from Zamzam water, sleeping, smiling, eating, praying, sitting, shaving hairs and ablutions,  reading the holy Quran and making duaa). The spatial resolutions are $1280 x 720$ pixels for images and $640 x 480$ pixels for videos and have lengths of $20$ seconds in average with $30$  frame per second  rates.\footnote{Thanks to HajjCoRE, Center of Research Excellence in Hajj~ and~ Umrah at~ UQU, ~an agency  for funding this work.}
\end{abstract}

\section{Introduction}\label{sec:intro}
In the last decade the field of visual recognition had an outstanding evolution from classifying instances of objects towards recognizing the classes of objects and scenes in natural images. Much of this progress has been sparked by the creation of realistic image datasets as well as by the new and  robust methods for image description and classification. We take inspiration from this progress and aim to transfer previous experience to the domain of video recognition and the recognition of human actions in particular during Hajj and Umrah seasons~\cite{Laptev2008}.

\goodbreak

Action recognition from video shares common problems with object recognition in static images. Both tasks have to deal with significant intra-class variations, background clutter and occlusions. In the context of object recognition in static images, these problems are surprisingly well handled by a bag-of-features representation \cite{Willamowski2004} combined with state-of-the-art machine learning techniques like support vector machines. It remains, however, an open question whether and how these results generalize to the recognition of realistic human actions, e.g., in feature films or personal videos.

\medskip

The Hajj and Umrah event recognition datasets are capable of recognizing a wide range of human actions during the Hajj and Umrah rituals (Tawaf, Sa'y, Arafat. etc) under different conditions. The main goal is to develop the  Hajj and Umrah event recognition datasets to solve all the following problems:

\begin{itemize}
\item Detecting injured, dead, and sleeping pilgrims inside Masjid El-Harram and around Kabaa.
\item Detecting abnormal pilgrims during Safa/Marwa and recognizing overcrowded areas.
\item Detecting abnormal human events in Makkah and in particular outside El-Harram.
\item Detecting people setting in sidewalks, allies, sub-roads, and stairs.
\item Detecting missing and found people or objects inside El-Harram and outside El-Harram.
\item Detecting and recognizing empty spaces inside and outside El-Harram.
\end{itemize}
\medskip

\begin{figure}[t]
\begin{center}$
\begin{array}{cc}
\includegraphics[width=3.5in, height=3.5in]{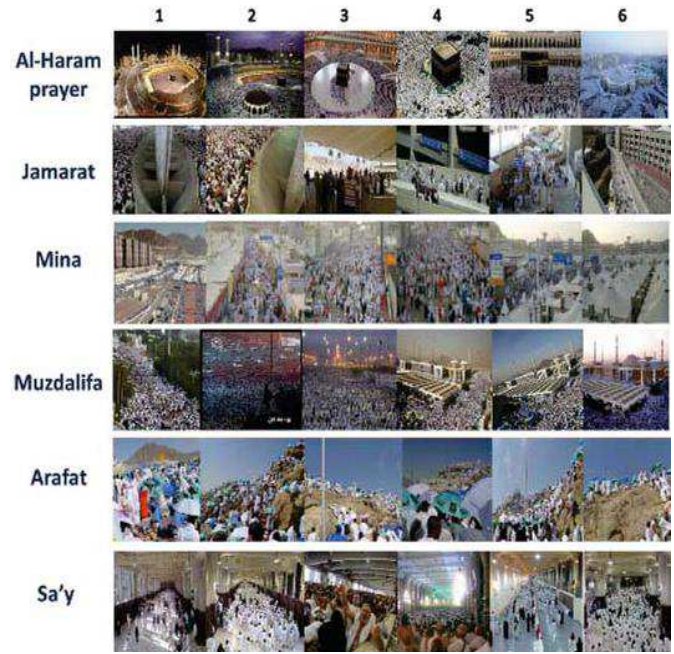}
\end{array}$
\end{center}
\caption{Hajj and Umrah Events Recognition Datasets. Various images are taken from different places representing Hajj and Umrah rituals.}
\label{fig:Dataset}
\end{figure}

\section{Hajj and Umrah rites activities and duties}

The Hajj is the fifth pillar of Islam and the Hajj is one of the largest pilgrimages in the world. The Hajj and Umrah are a demonstration of the solidarity of the Muslim people, and their submission to Allah. Many pilgrims come simultaneously converge on Makkah to do Hajj and Umrah and perform a series of rituals: Each pilgrim walks seven times around the Kaaba counter-clockwise, then runs back and forth between the Al-Safa and Al-Marwah hills, also the pilgrims always drink from the Zamzam Well, however the pilgrims go to the plains of mount of Arafat to stand in vigil, and throws stones in a ritual Stoning of the Devil, see for example~\cite{Sabiq1992}. The pilgrims then shave their heads, perform a ritual of animal sacrifice, and celebrate the three day global festival of Eid al-Adha, see the Hajj and Umrah events' recognition in Fig.~\ref{fig:Dataset}.

There are six hajj and umrah ritual events, which are modeling during Hajj and Umrah as seen in  Fig.~\ref{fig:AllRites}. These are circling in Tawaf, performing Sa'y between Safa and Marwa, standing on the mount of Arafat, staying overnight in Muzdalifah, staying three nights in Mina, and throwing  Jamarat ~\cite{Shirazi2003}. The models defined for this study are described below:

\begin{figure}[t]
\begin{center}$
\begin{array}{cc}
\includegraphics[width=3.5in, height=2.3in]{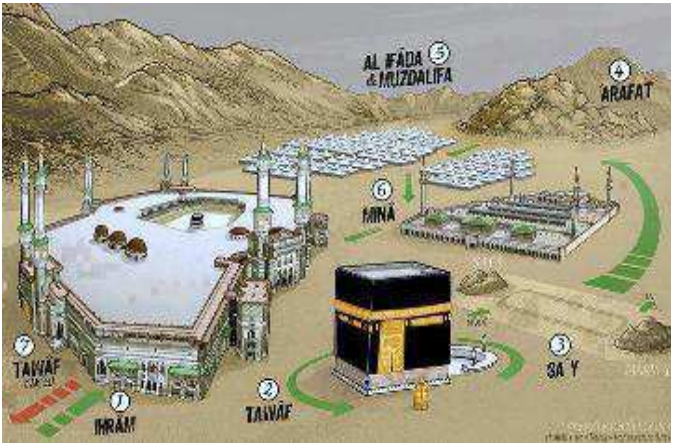}
\end{array}$
\end{center}
\caption{All Hajj and Umrah ritual locations}
\label{fig:AllRites}
\end{figure}

\begin{enumerate}

\item{\textbf{Pilgrims Tawaf:}}

The pilgrims Tawaf around Kaaba is one of the most Islamic rituals. During the Hajj and Umrah, Muslims have to circumambulate the Kaaba (the most sacred site in Islam, and the most sacred place on earth) seven times, in a counterclockwise direction~\cite{Sabiq1992}~\cite{Al-Islamwebsite}, see Fig.~\ref{fig:Tawaf}.

\goodbreak

\begin{figure}[h]
\begin{center}$
\begin{array}{cc}
\includegraphics[width=1.5in, height=1.1in]{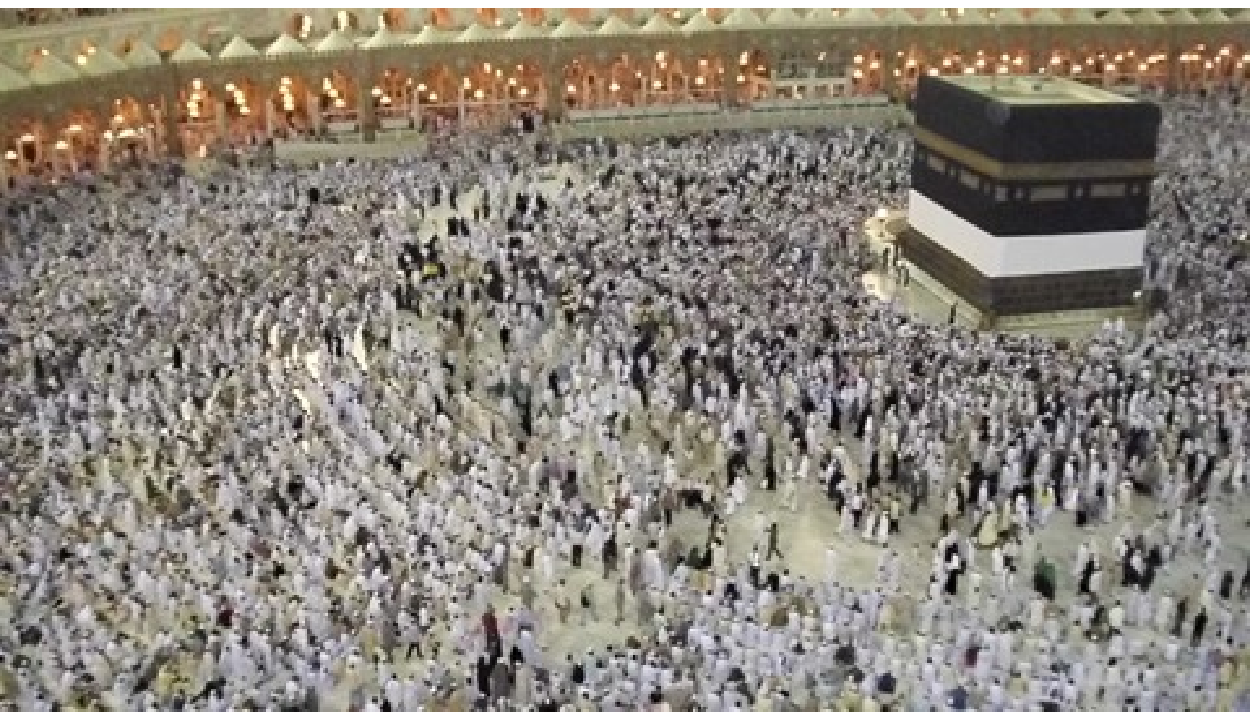} &
\includegraphics[width=1.5in, height=1.1in]{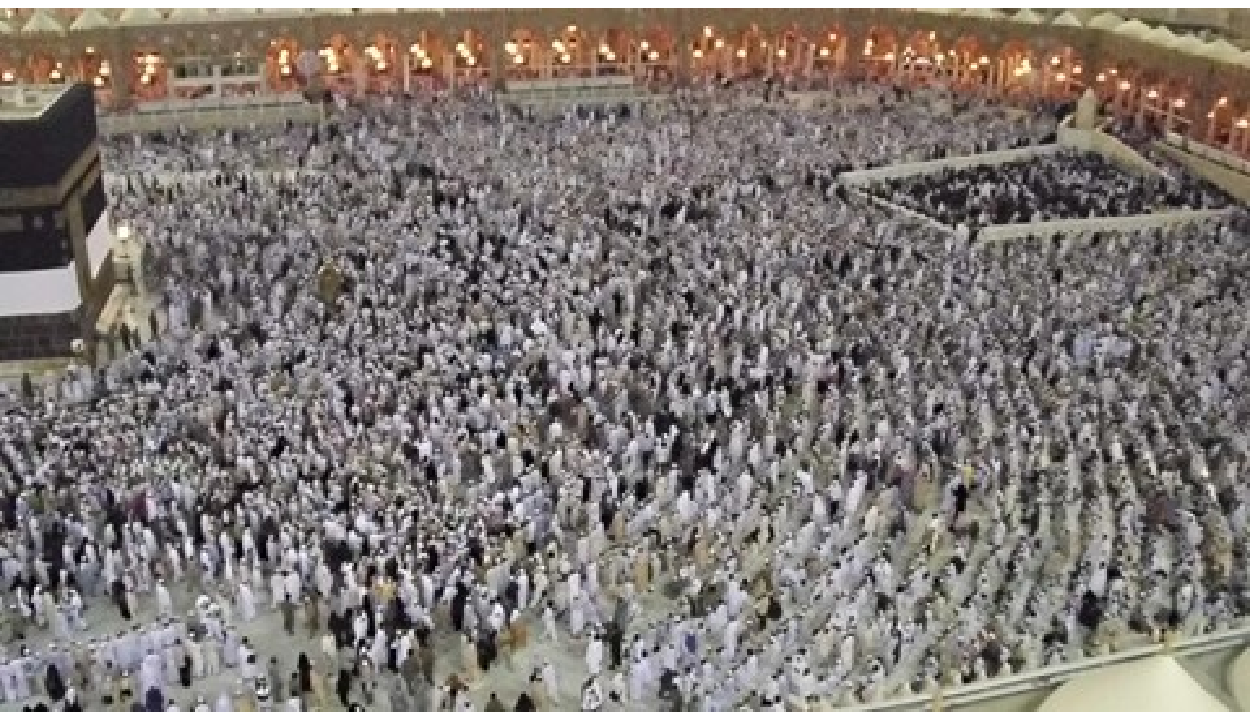}
\end{array}$
\end{center}
\caption{Examples of the pilgrims circling in Tawaf}
\label{fig:Tawaf}
\end{figure}

\item{\textbf{Sa'y between Safa and Marwa:}}
Pilgrims, whether they are performing Hajj or Umrah, perform Sa'y after Tawaf. Sa'y means endeavoring or making effort. For Hajj, this is held to commemorate Hagar's running between Safa and Marwa seven times in order to find water for her son, Ishmael, whom she was still breast-feeding~\cite{Sabiq1992}~\cite{Al-Islamwebsite}, see Fig.~\ref{fig:Safa and Marwa}.\\

\begin{figure}[h]
\begin{center}$
\begin{array}{cc}
\includegraphics[width=1.5in, height=1.1in]{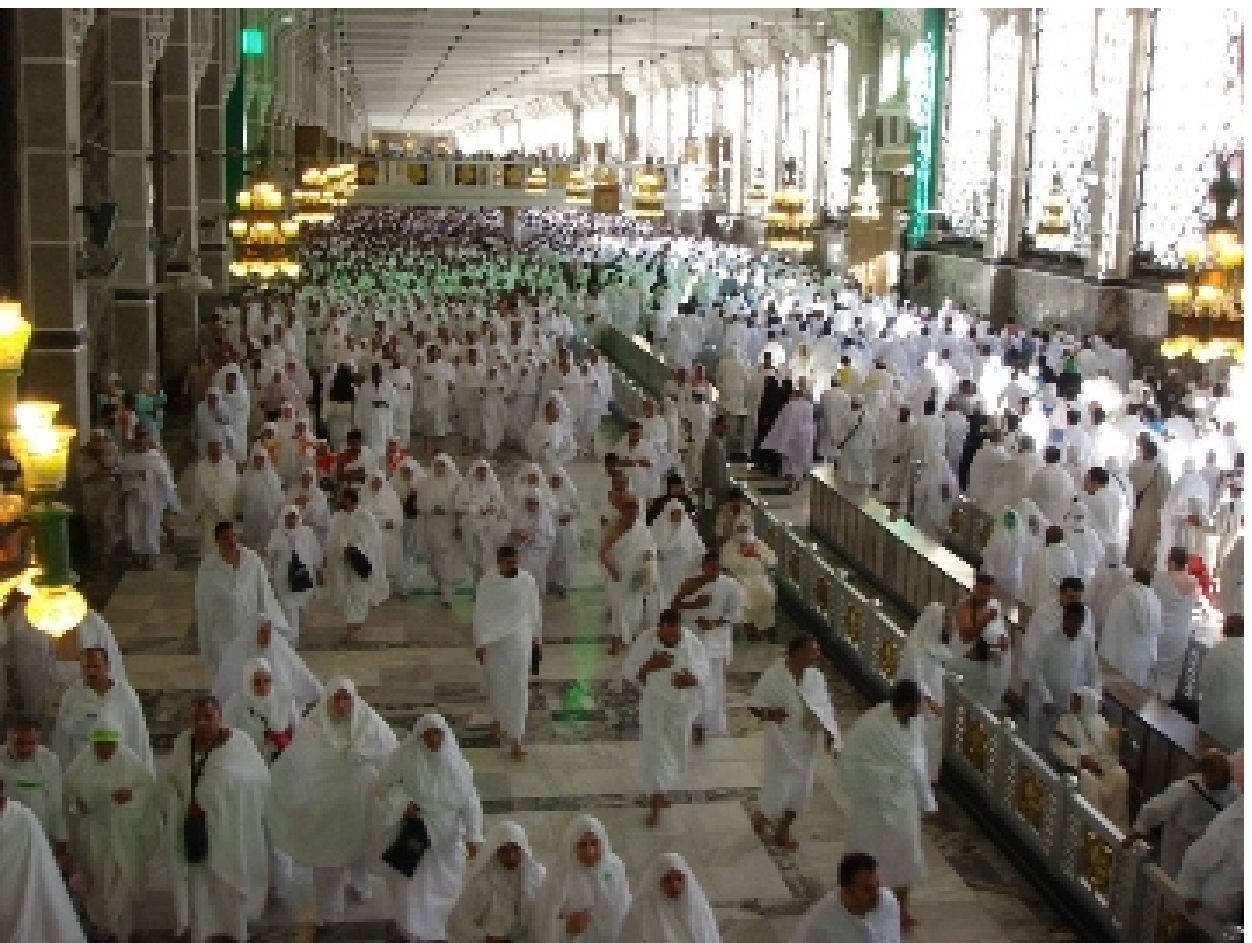} &
\includegraphics[width=1.5in, height=1.1in]{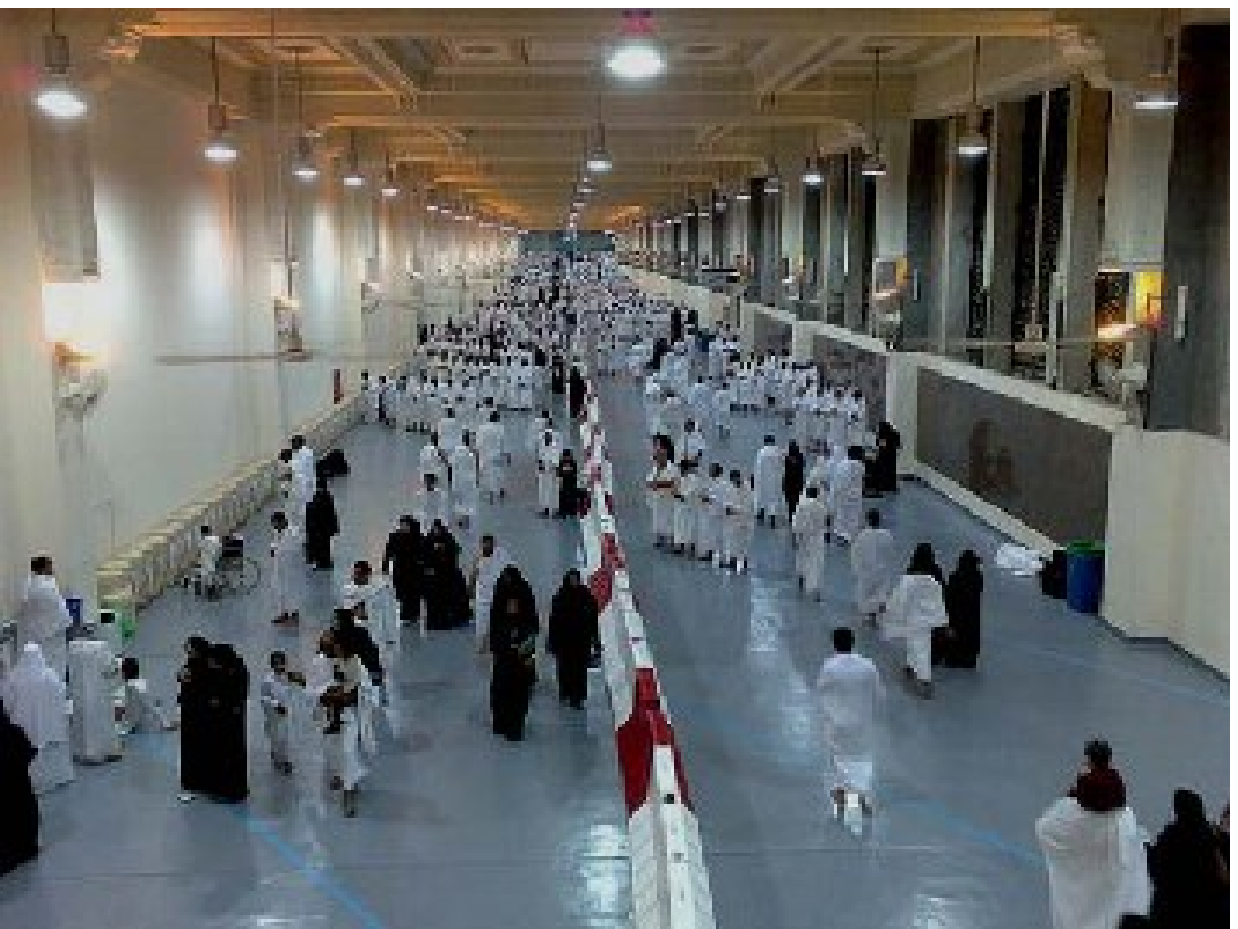}
\end{array}$
\end{center}
\caption{Examples of Sa'y between Safa and Marwa}
\label{fig:Safa and Marwa}
\end{figure}

\item{\textbf{Standing on the mount of Arafat:}}
The plain and mount of Arafat are located in the south-east side of Makkah approximately $12$ kilometers away of El-Harram,  Saudi Arabia, about three million pilgrims congregated to perform the most important rite of the Hajj, or the pilgrimage. This rite is significant because it is on the mount of Mercy that the Prophet Muhammad gave his final sermon. Many pilgrims climb the hill and try to touch the pillar that marks this place. After Arafat, pilgrims will move to Muzdalifah to complete the remaining rites of the pilgrimage~\cite{Sabiq1992}~\cite{Al-Islamwebsite}, see Fig.~\ref{fig:Arafat}.

\begin{figure}[h]
\begin{center}$
\begin{array}{cc}
\includegraphics[width=1.5in, height=1.1in]{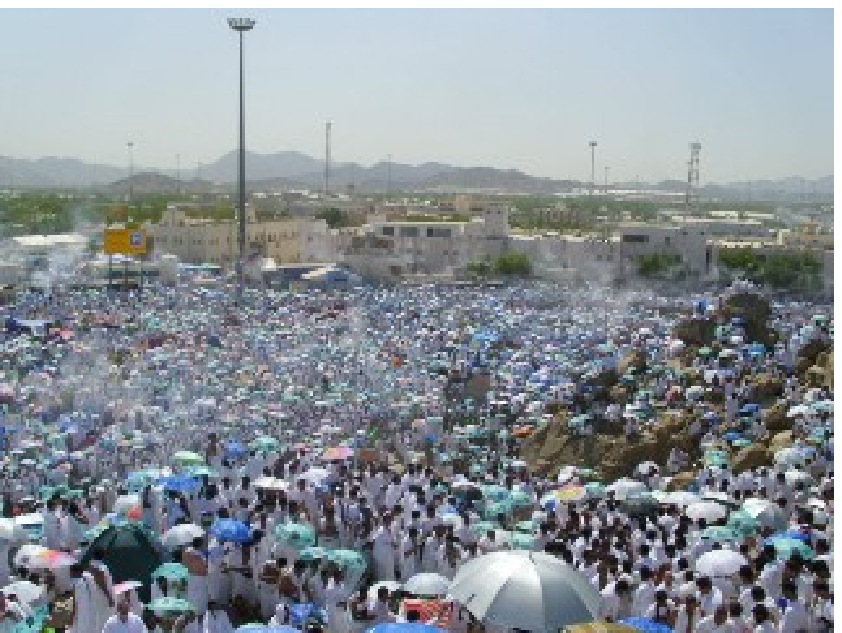} &
\includegraphics[width=1.5in, height=1.1in]{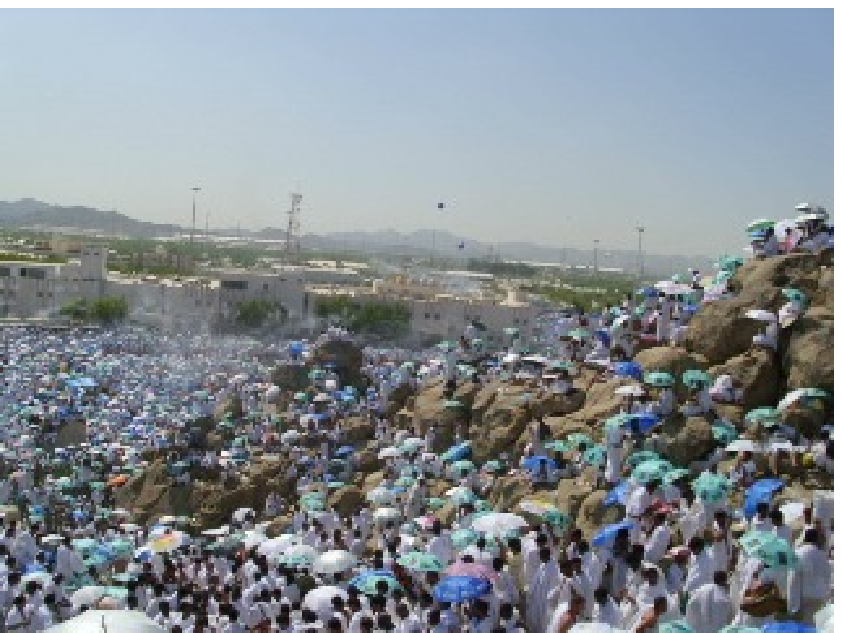}
\end{array}$
\end{center}
\caption{Examples of standing on the mount of Arafat}
\label{fig:Arafat}
\end{figure}

\item{\textbf{Staying overnight in Muzdalifah:}}
Staying in Muzdalifah is obligatory upon the one performing the Hajj to spend the tenth $(10^{th})$ of Dhul-Hijjah until the time of Fajr prayer~\cite{Sabiq1992}~\cite{Al-Islamwebsite}, see Fig.~\ref{fig:Muzdalifah}.

\begin{figure}[h]
\begin{center}$
\begin{array}{cc}
\includegraphics[width=1.5in, height=1.1in]{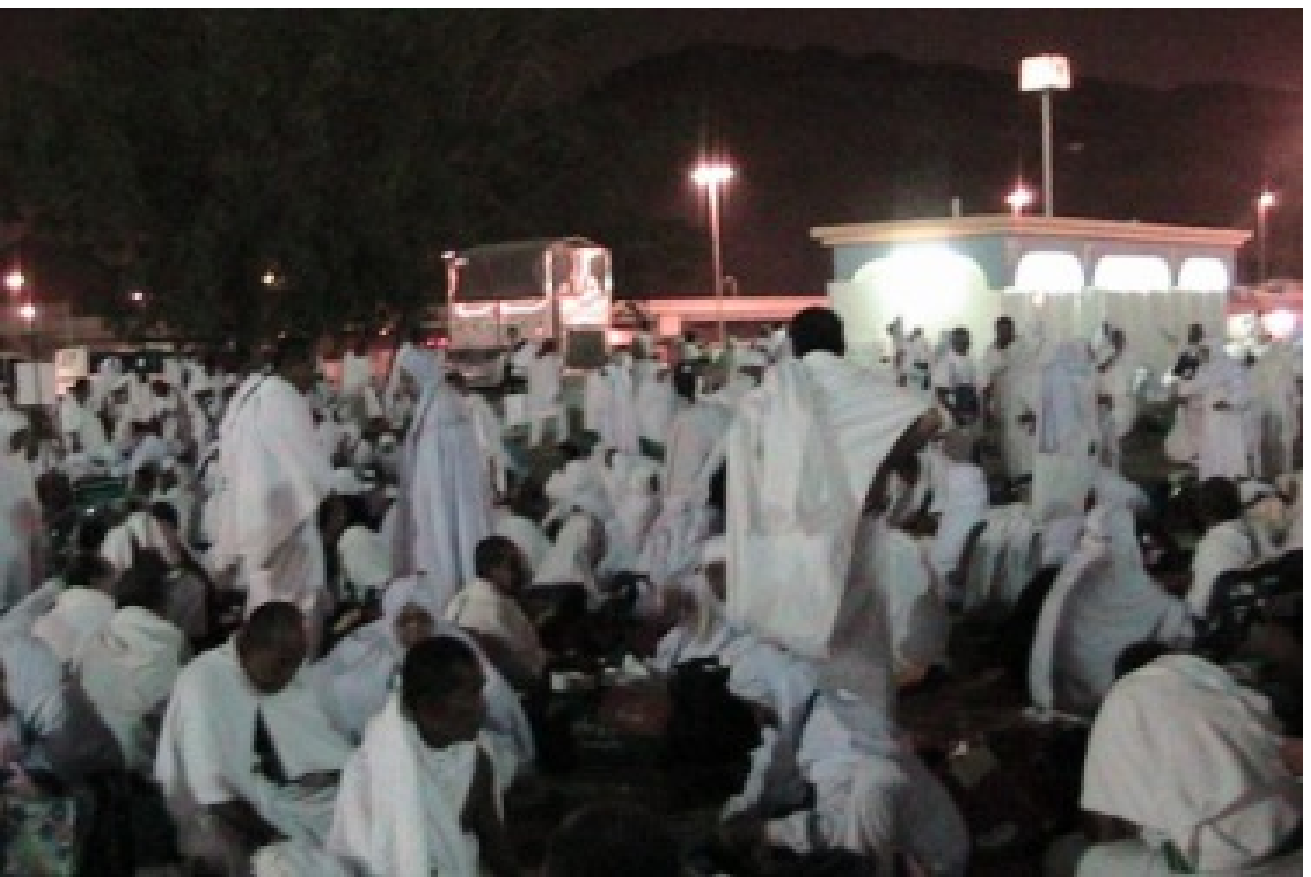} &
\includegraphics[width=1.5in, height=1.1in]{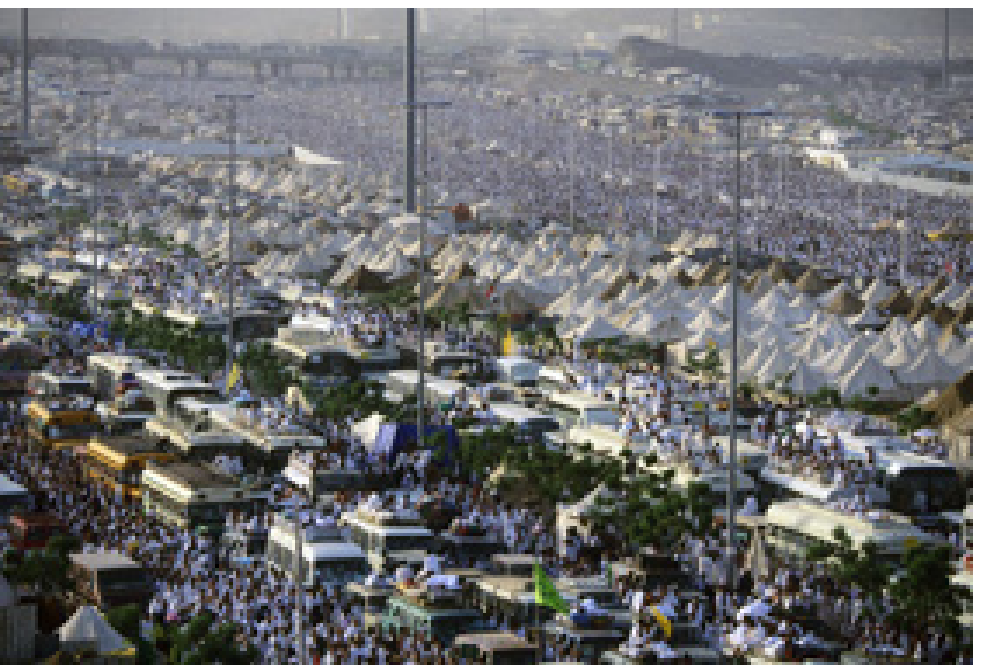}
\end{array}$
\end{center}
\caption{Examples of staying overnight in Muzdalifah}
\label{fig:Muzdalifah}
\end{figure}

\item{\textbf{Staying overnight in Mina:}}
Mina, seven kilometers east-south of the Masjid El-Harram is where Hajj pilgrims sleep overnight on the $8^{th}$, $11^{th}$, $12^{th}$ (and some even on the $13^{th}$) of Dhul Hijjah. It contains the Jamarat, the three stone pillars which are pelted by pilgrims as part of the rituals of Hajj~\cite{Sabiq1992}~\cite{Al-Islamwebsite}, see Fig.~\ref{fig:Mina}.

\begin{figure}[h]
\begin{center}$
\begin{array}{cc}
\includegraphics[width=1.5in, height=1.1in]{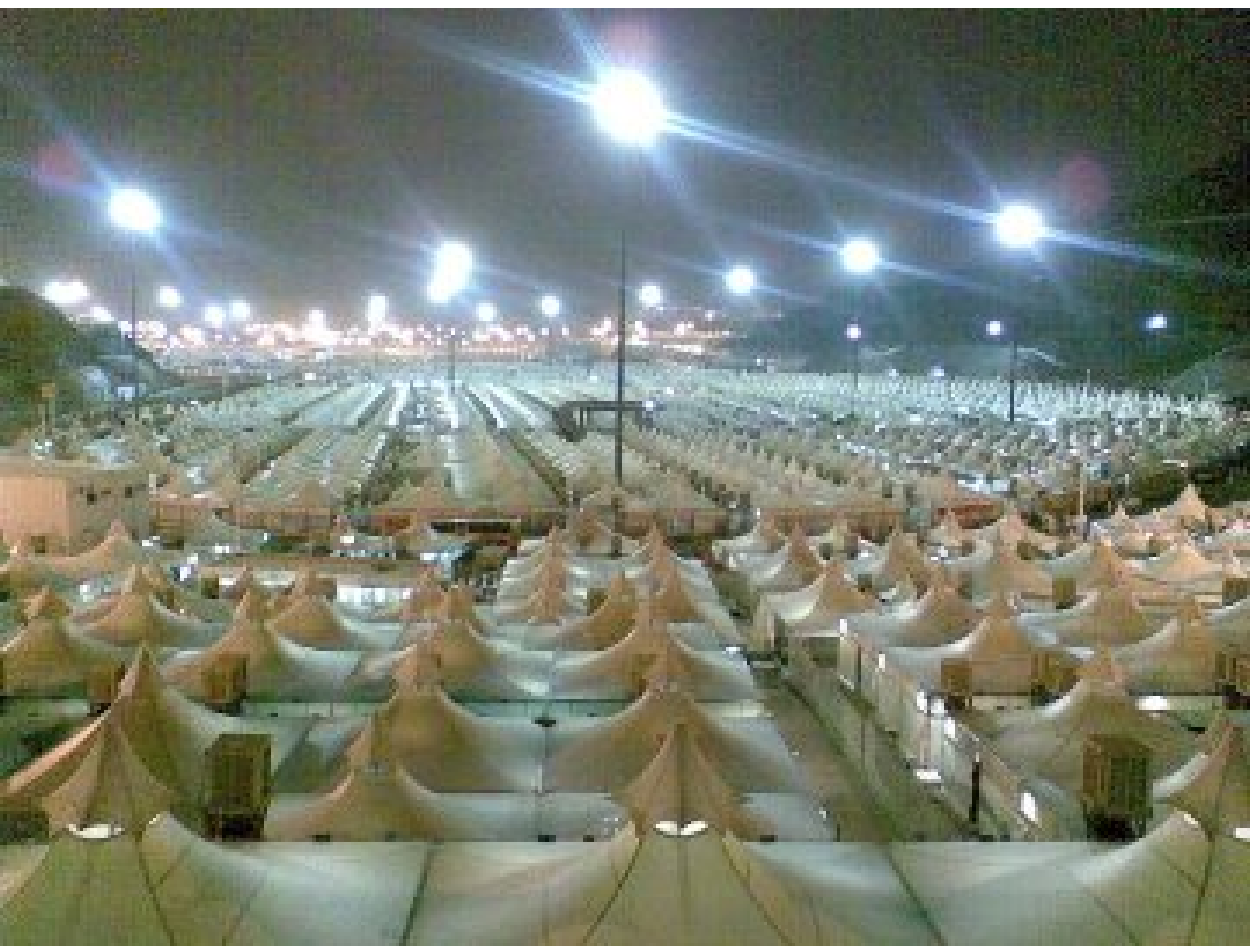} &
\includegraphics[width=1.5in, height=1.1in]{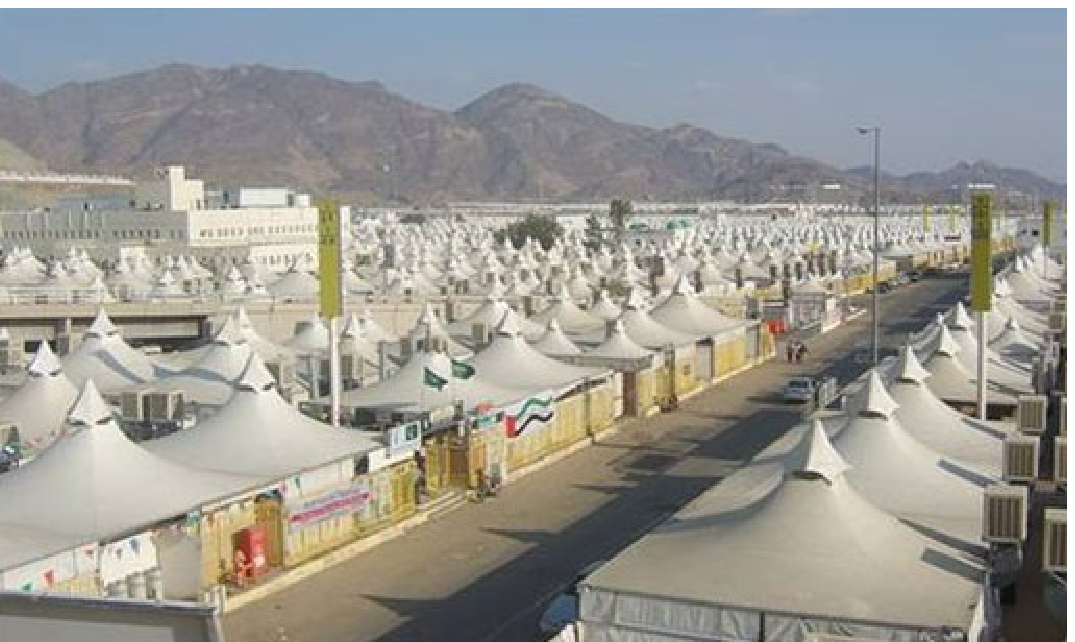}
\end{array}$
\end{center}
\caption{Examples of staying overnight in Mina}
\label{fig:Mina}
\end{figure}

\item{\textbf{Throwing Jamarat:}}
The pilgrim who throws Jamarat before Zawal on the $11^{th}$ day and the following days will have to throw the pebbles again after Zawal if the days of throwing the pebbles have not yet expired~\cite{Sabiq1992}~\cite{Al-Islamwebsite}, see Fig.~\ref{fig:Jamarat}.\\

\begin{figure}[h]
\begin{center}$
\begin{array}{cc}
\includegraphics[width=1.5in, height=1.1in]{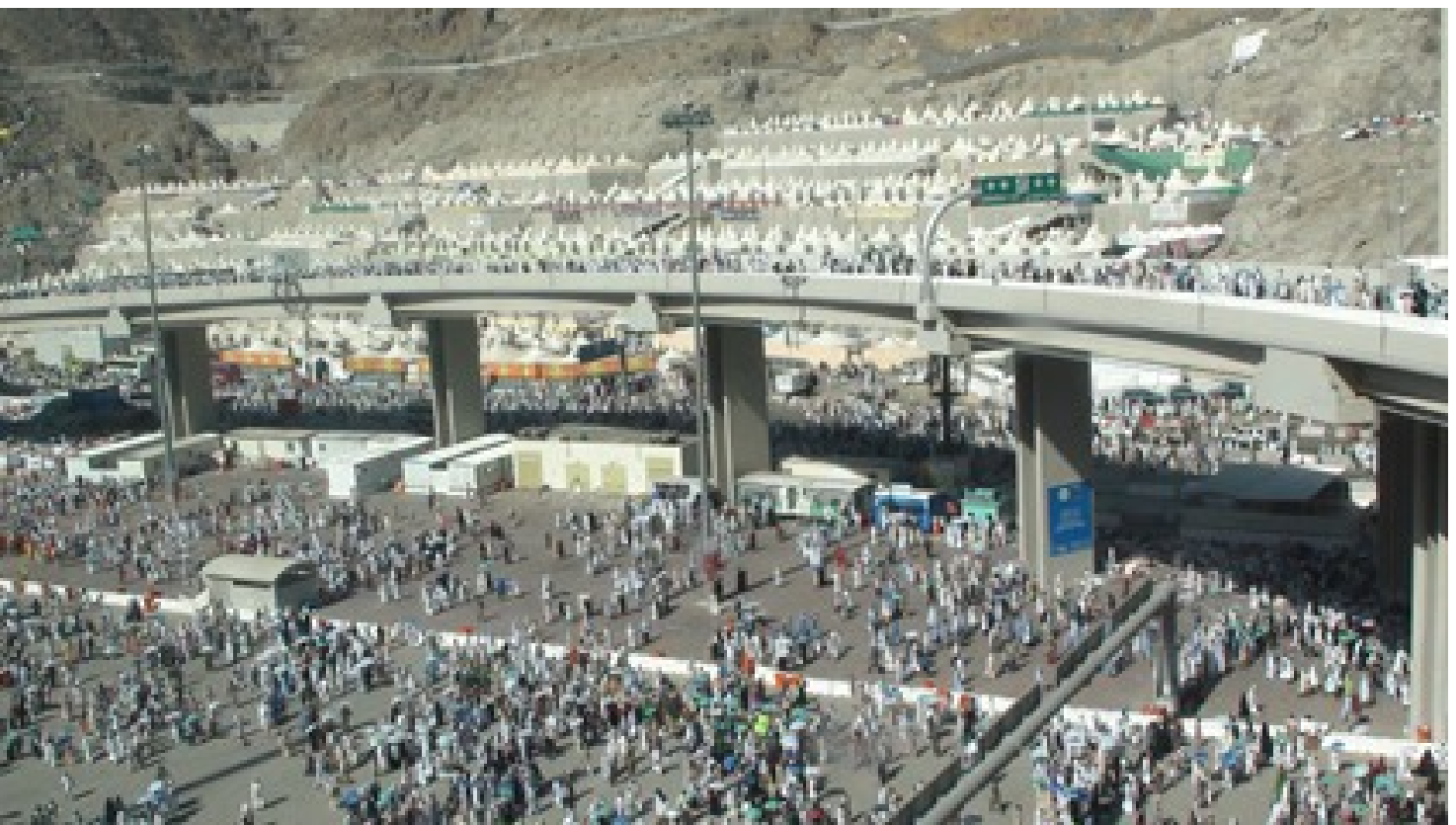} &
\includegraphics[width=1.5in, height=1.1in]{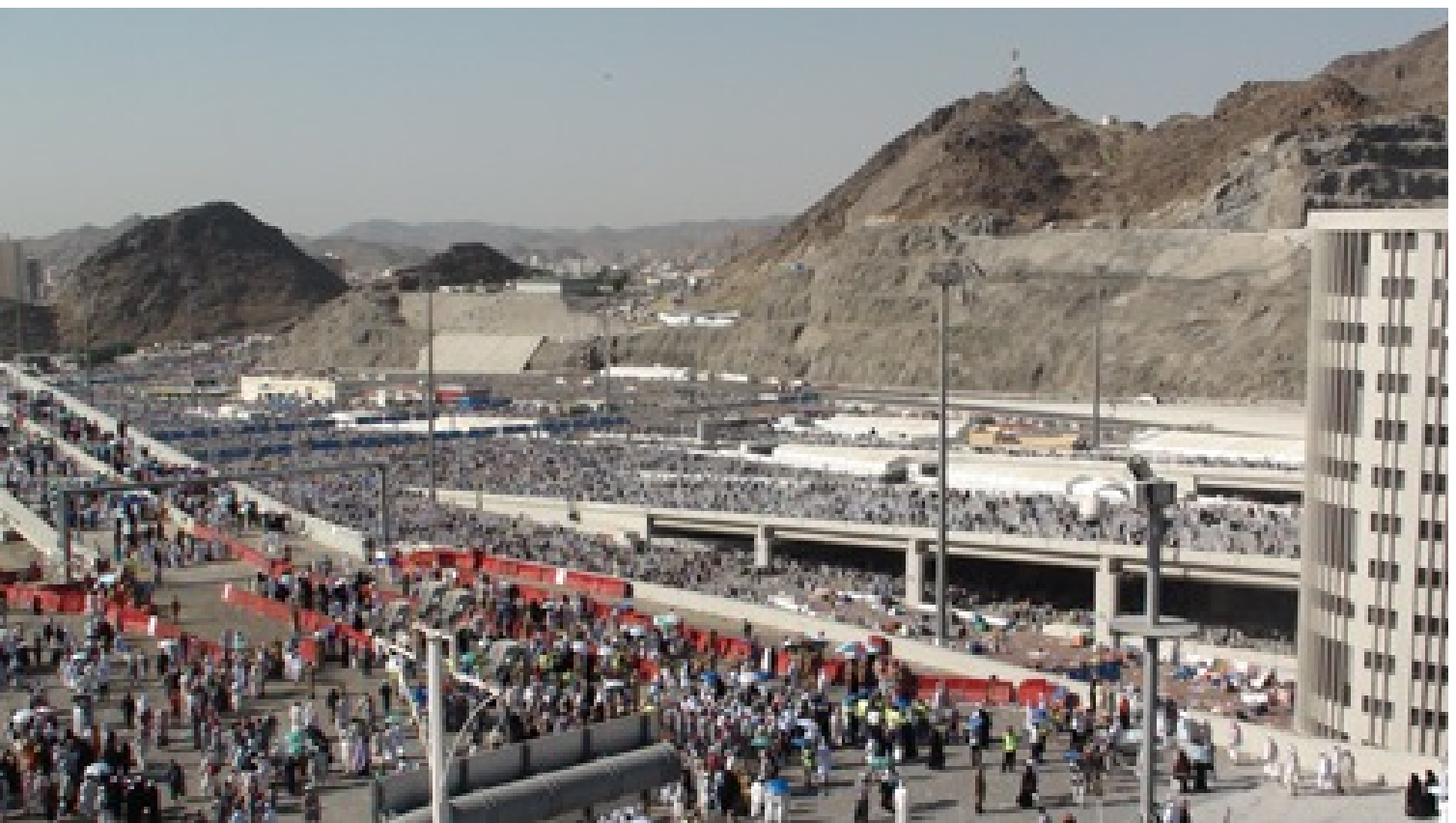}
\end{array}$
\end{center}
\caption{Examples of throwing Jamarat}
\label{fig:Jamarat}
\end{figure}

\end{enumerate}

\section{Hajj and Umrah Pilgrim Event Classifications}

Nine pilgrim events will be modeled during Hajj and Umrah rituals, as shown in Fig.~\ref{fig:Dataset2}. They are drinking from Zamzam water, walking pilgrims, smiling pilgrims, sleeping pilgrims, sitting pilgrims, eating pilgrims, praying  pilgrims, shaving hairs and ablution pilgrims, and reading quran and making duaa. The models defined for this study are described below:

\medskip

\begin{figure}[h]
\begin{center}$
\begin{array}{cc}
\includegraphics[width=3.5in, height=3.0in]{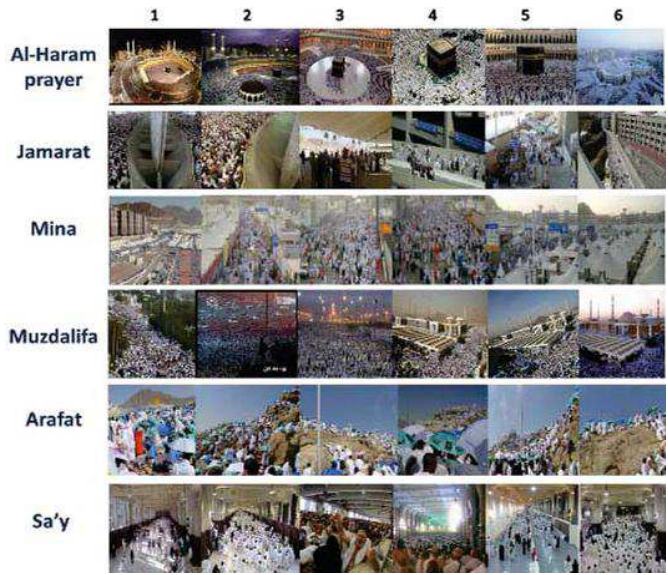}
\end{array}$
\end{center}
\caption{Hajj and Umrah Pilgrim Events Recognition Datasets. Various images are taken from different places representing Hajj and Umrah rituals.}
\label{fig:Dataset2}
\end{figure}

\begin{enumerate}

\item{\textbf{Walking pilgrims:}}
Walking during Hajj corresponds to a significant number of pilgrims moving at a low speed. The pilgrims are walking in  all rites for long distances in huge numbers, which resulting in overcrowding and suffocation, see Fig.~\ref{fig:Walking}.\\

\begin{figure}[h]
\begin{center}$
\begin{array}{cc}
\includegraphics[width=1.5in, height=1.1in]{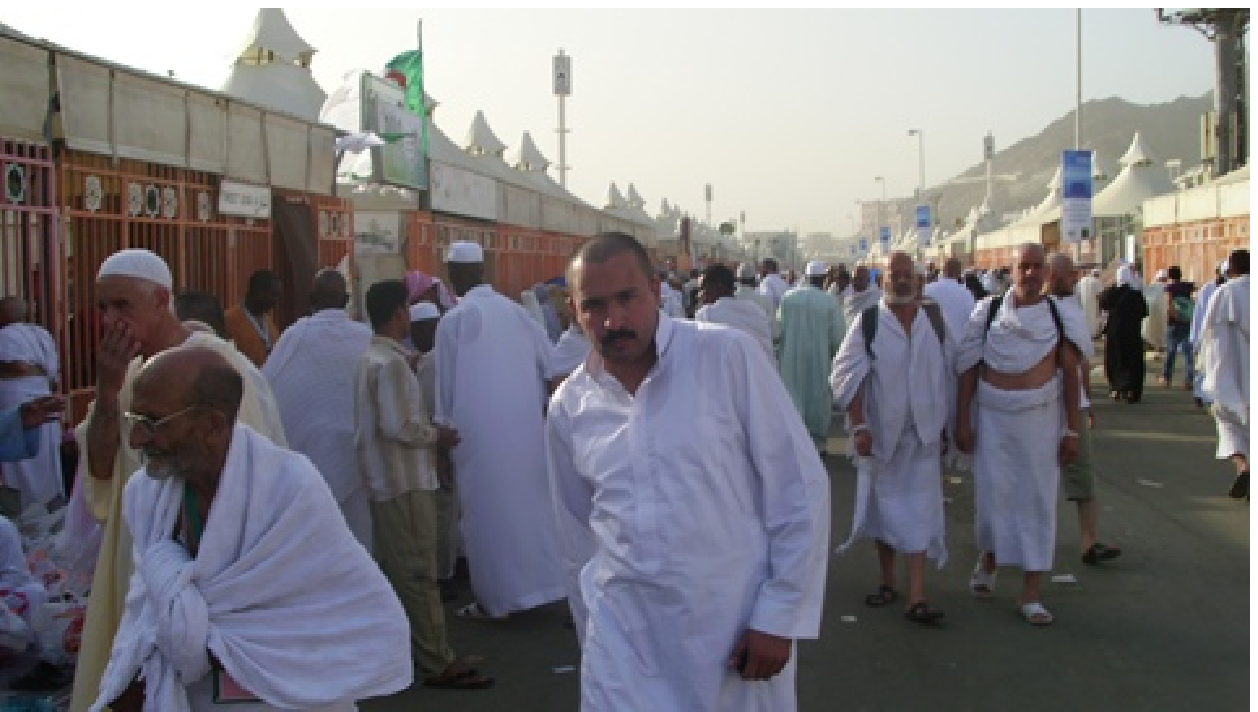} &
\includegraphics[width=1.5in, height=1.1in]{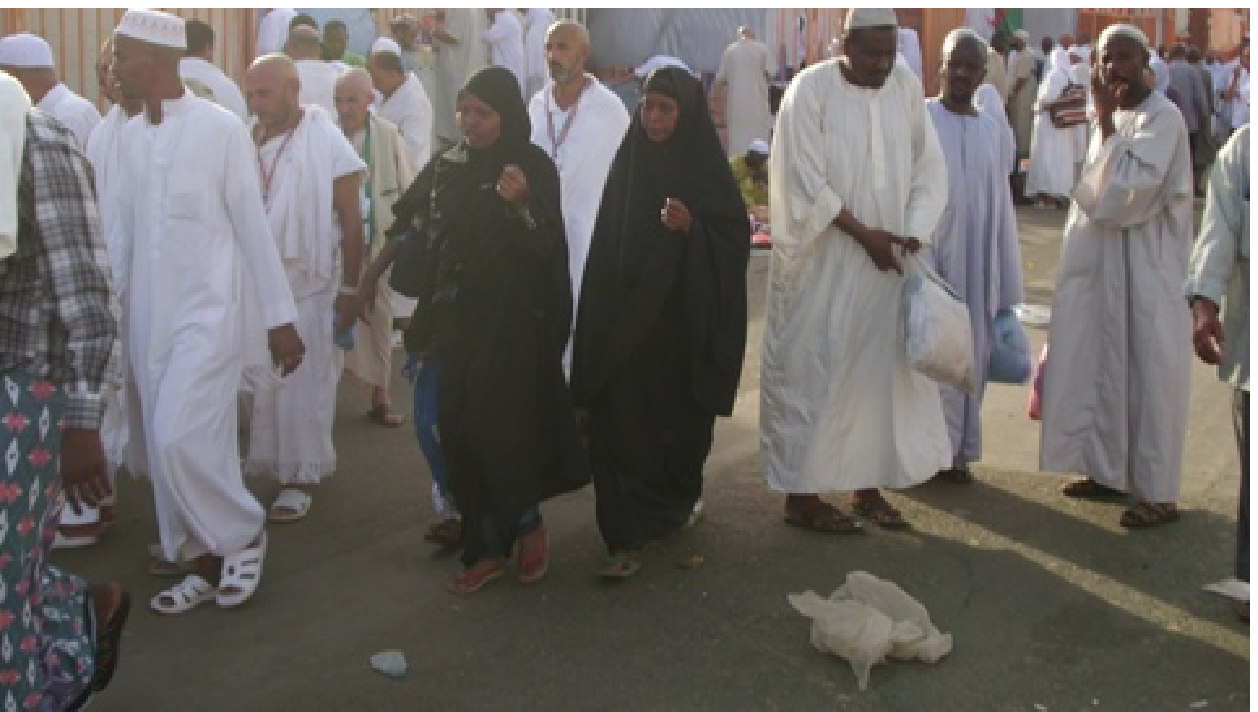}
\end{array}$
\end{center}
\caption{Examples of the walking pilgrims}
\label{fig:Walking}
\end{figure}

\item{\textbf{Drinking from Zamzam water:}}
The Zamzam well is located in Makkah, which is the heart of the Hajj pilgrimage. Standing only a few meters east of the Kabaah, the well is 35 meters deep and topped by an elegant dome. In last decades, scientists have collected samples of Zamzam water and they  have found certain peculiarities that make water healthier, like a higher level of calcium, see Fig.~\ref{fig:Zamzam}.

\begin{figure}[h]
\begin{center}$
\begin{array}{cc}
\includegraphics[width=1.5in, height=1.1in]{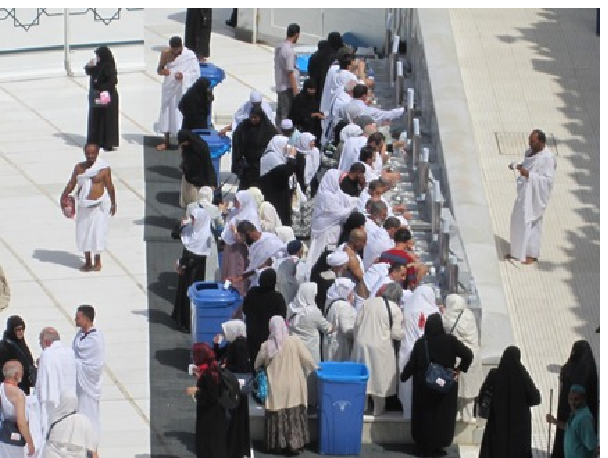} &
\includegraphics[width=1.5in, height=1.1in]{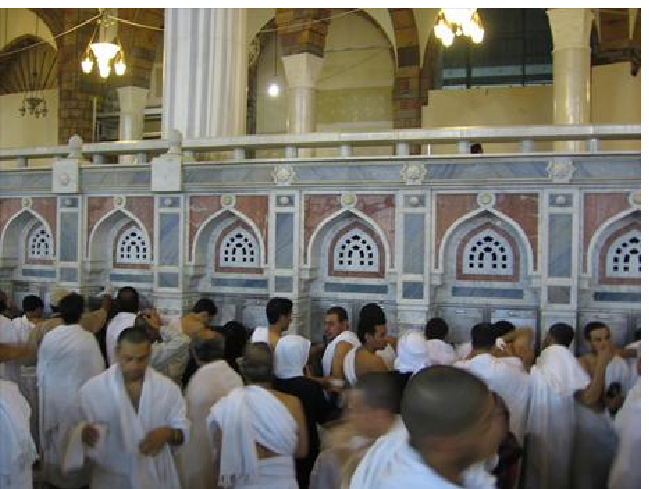}
\end{array}$
\end{center}
\caption{Examples of drinking from Zamzam water}
\label{fig:Zamzam}
\end{figure}

\item{\textbf{Sleeping pilgrims:}}
This event will recognize the Pilgrims sleeping from severity of fatigue and exhaustion during Hajj and Umrah rites. Therefore, some of them may resort to sleep in the streets, putting their lives at risk,  see Fig.~\ref{fig:Sleeping}.

\begin{figure}[h]
\begin{center}$
\begin{array}{cc}
\includegraphics[width=1.5in, height=1.1in]{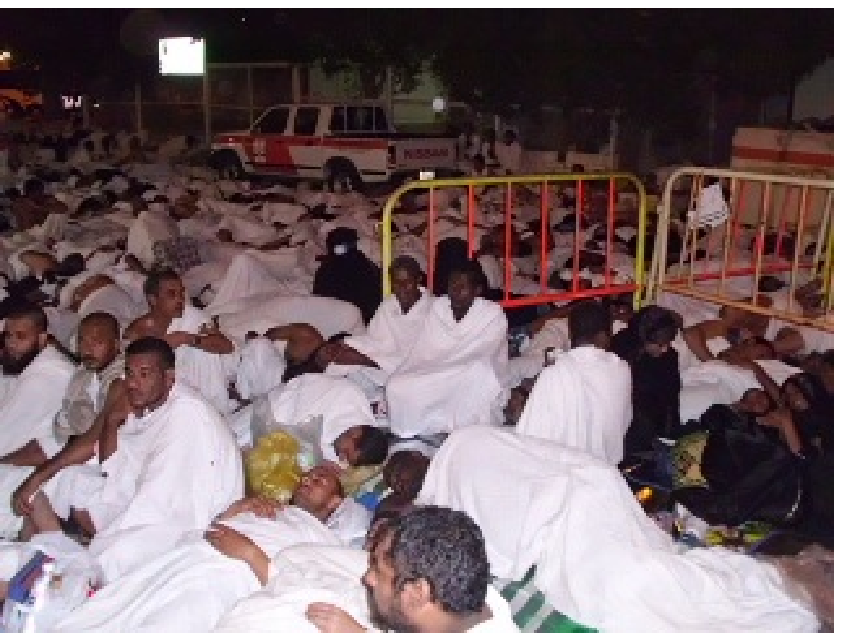} &
\includegraphics[width=1.5in, height=1.1in]{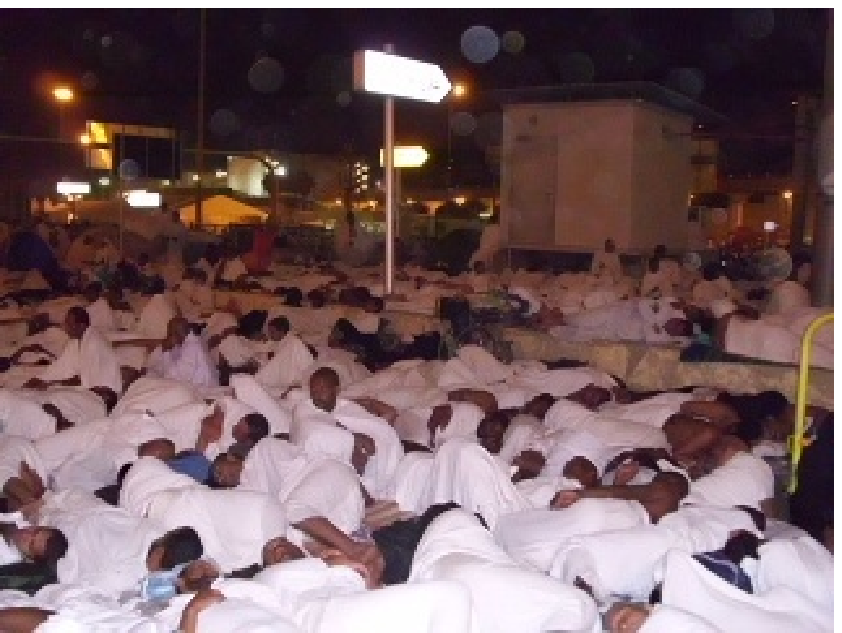}
\end{array}$
\end{center}
\caption{Examples of the sleeping pilgrims}
\label{fig:Sleeping}
\end{figure}

\item{\textbf{Eating pilgrims:}}
This event will recognize the pilgrims eating during the Hajj and Umrah in the camps and places to spend the night in Mina and Muzdalifah,  see Fig.~\ref{fig:Eating}.

\begin{figure}[h]
\begin{center}$
\begin{array}{cc}
\includegraphics[width=1.5in, height=1.1in]{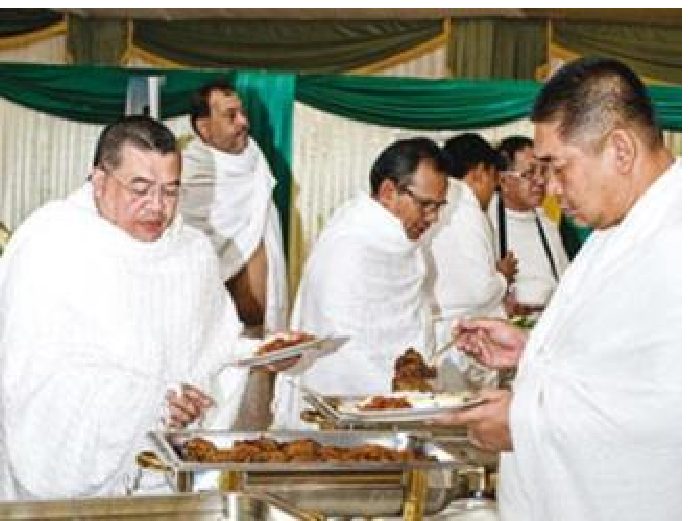} &
\includegraphics[width=1.5in, height=1.1in]{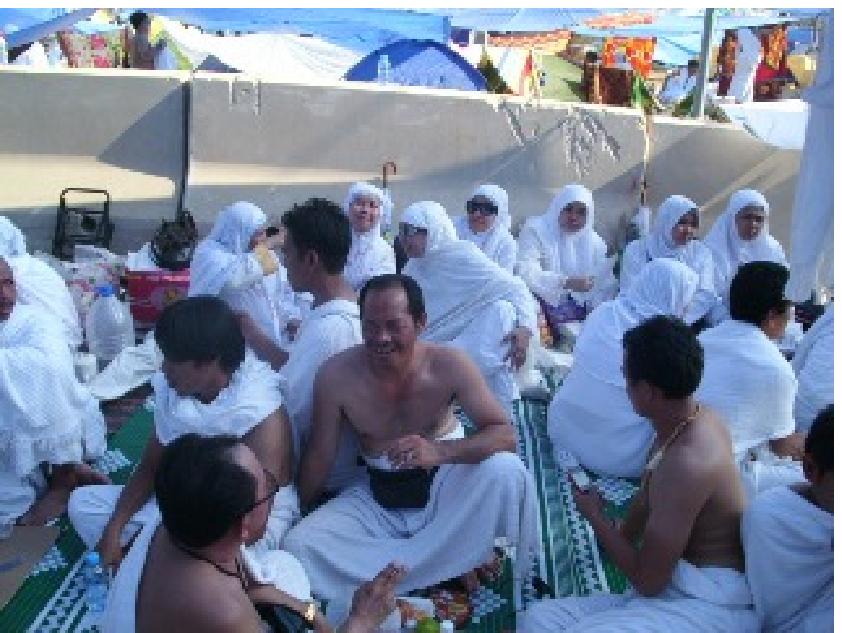}
\end{array}$
\end{center}
\caption{Examples of the eating pilgrims}
\label{fig:Eating}
\end{figure}

\item{\textbf{Al-Haram Prayer pilgrims:}}
This event will recognize pilgrims performing most or all five daily prayers in the most beautiful and purest part of the earth in the Holy Mosque in Makkah, see Fig.~\ref{fig:Prayer}.

\begin{figure}[h]
\begin{center}$
\begin{array}{cc}
\includegraphics[width=1.5in, height=1.1in]{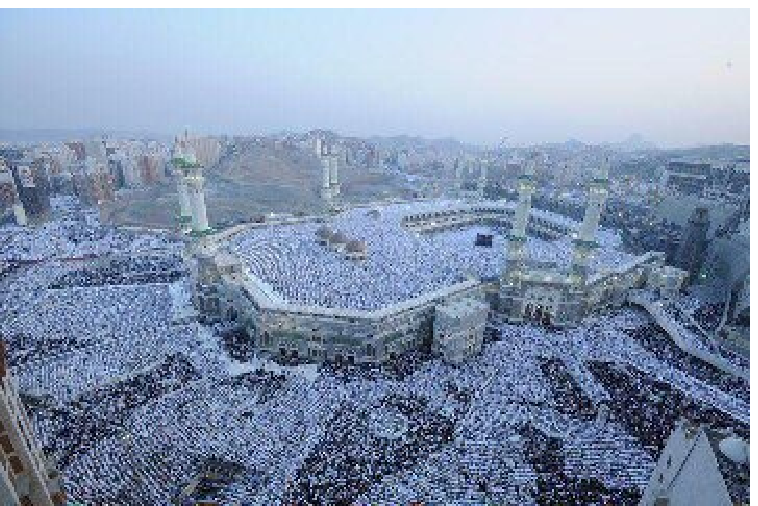} &
\includegraphics[width=1.5in, height=1.1in]{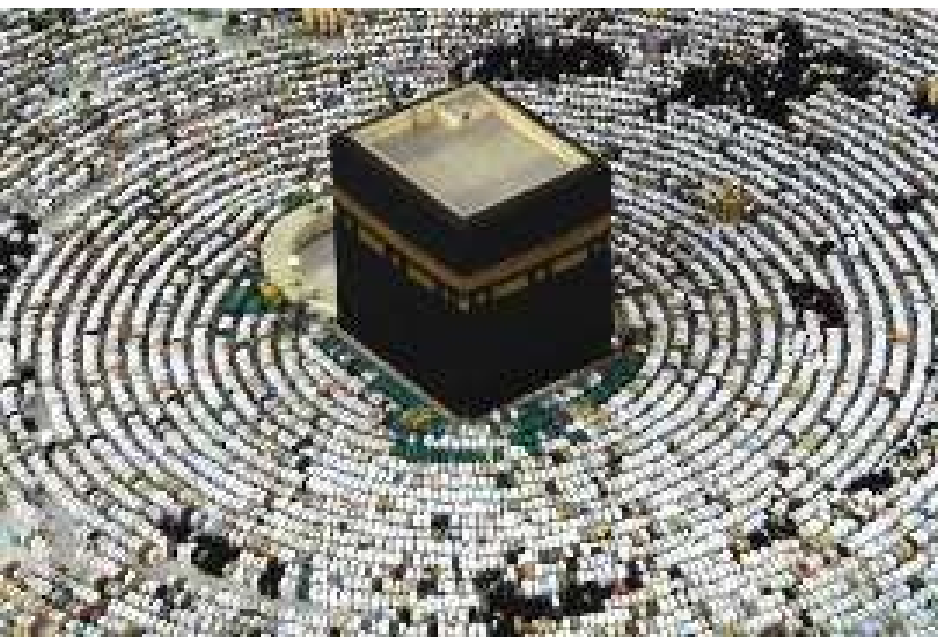}
\end{array}$
\end{center}
\caption{Examples of the praying pilgrims}
\label{fig:Prayer}
\end{figure}


\item{\textbf{Smiling pilgrims:}}
The smiling pilgrims is coming from their great happiness because they are in the best place where the survival of psychological serenity and peace of mind because they are with God (ALLAH), see Fig.~\ref{fig:Smiling}.

\begin{figure}[h]
\begin{center}$
\begin{array}{cc}
\includegraphics[width=1.5in, height=1.1in]{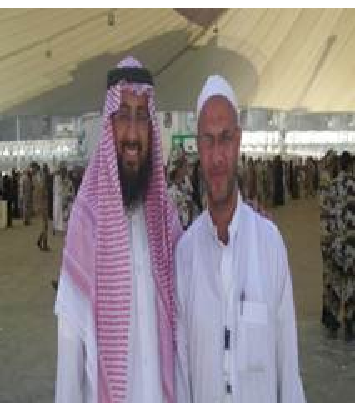} &
\includegraphics[width=1.5in, height=1.1in]{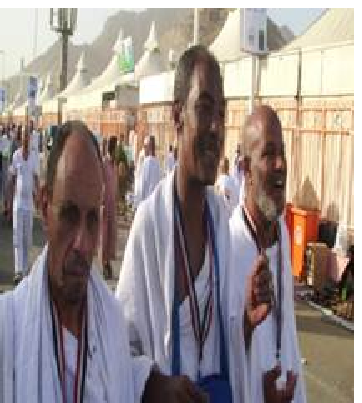}
\end{array}$
\end{center}
\caption{Examples of smiling pilgrims}
\label{fig:Smiling}
\end{figure}

\item{\textbf{Sitting pilgrims:}}

This event will recognize pilgrims siting during and after the performance of rituals, from the severity of fatigue. There are many places where having rest is possible, but some pilgrims may sit down and resting in the streets, which could lead to congestion in some of these areas from the intensity of the crowds, see Fig.~\ref{fig:Sitting}.

\begin{figure}[h]
\begin{center}$
\begin{array}{cc}
\includegraphics[width=1.5in, height=1.1in]{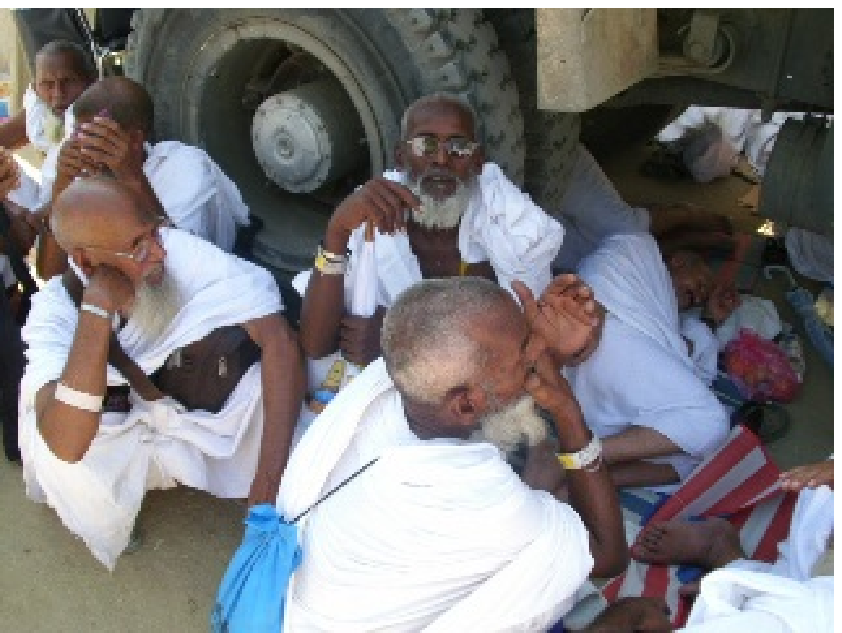} &
\includegraphics[width=1.5in, height=1.1in]{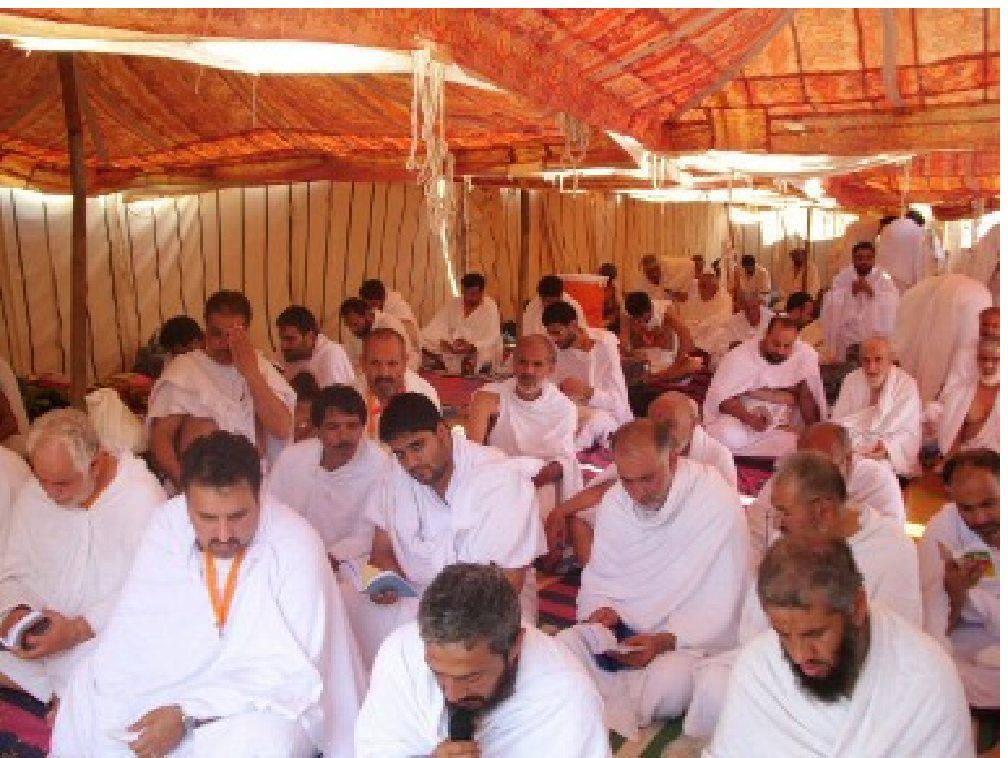}
\end{array}$
\end{center}
\caption{Examples of the sitting pilgrims}
\label{fig:Sitting}
\end{figure}

\item{\textbf{Ablutions and shaving hairs:}}
This event will recognize ablution, which is the cornerstone of the prayer and can't valid without it,  as well as it recognizes pilgrims shaving their hair during the Hajj or Umrah, see Fig.~\ref{fig:Ablutions}.\\

\begin{figure}[h]
\begin{center}$
\begin{array}{cc}
\includegraphics[width=1.5in, height=1.1in]{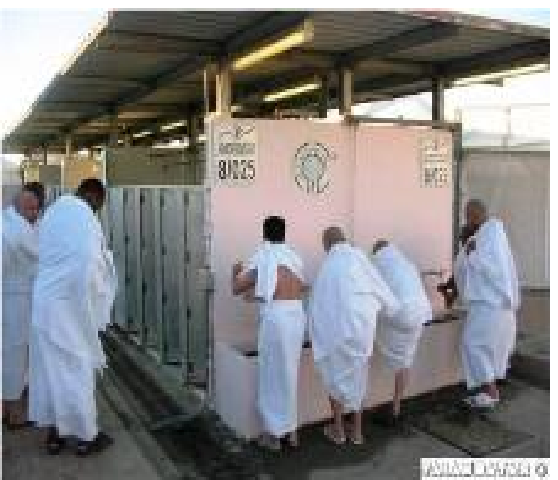} &
\includegraphics[width=1.5in, height=1.1in]{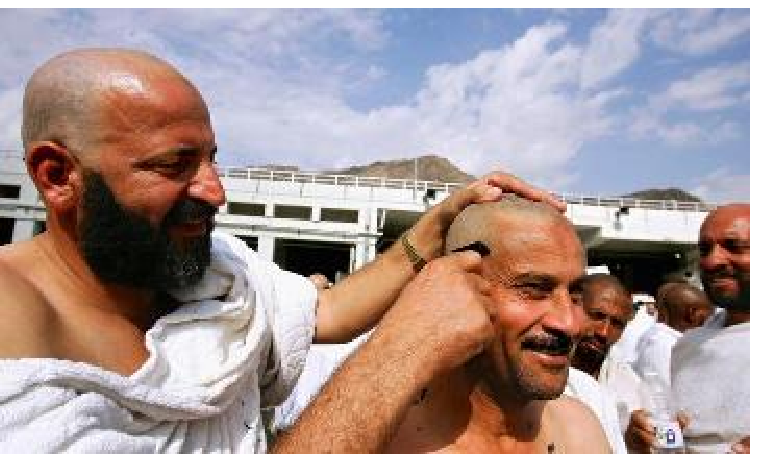}
\end{array}$
\end{center}
\caption{Ablutions and shaving hairs}
\label{fig:Ablutions}
\end{figure}

\item{\textbf{Reading quran and making duaa:}}
Reading the Holy Quran is one of the most important worship by Muslims, especially in the month of Ramadan.   the pious Muslims read regularly the Holy Quran, see Fig.~\ref{fig:Reading}.

\begin{figure}[h]
\begin{center}$
\begin{array}{cc}
\includegraphics[width=1.5in, height=1.1in]{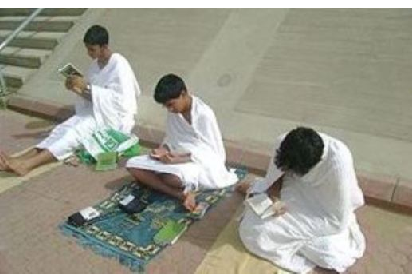} &
\includegraphics[width=1.5in, height=1.1in]{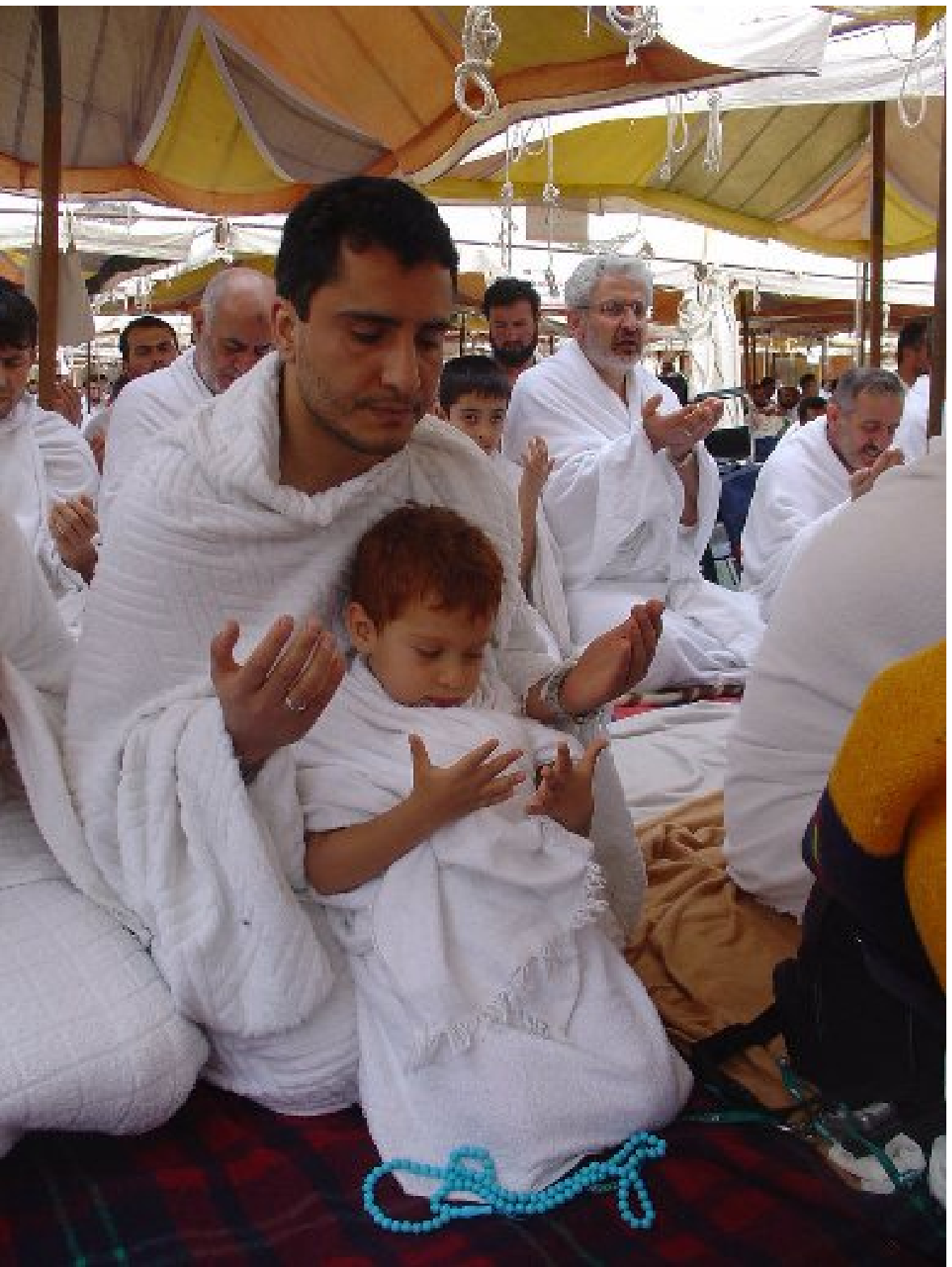}
\end{array}$
\end{center}
\caption{Reading the holy Quran and making duaa}
\label{fig:Reading}
\end{figure}

\end{enumerate}

\section{Discussion and Conclusion}
Up to our knowledge, the proposed human event recognition datasets are the first event recognition datasets to be developed to observe  and model the Hajj and Umrah activities in Makkah. We plan to develop algorithms for event recognition, videos extractions, and classifications based on these  developed datasets.

Other datasets of human events can be briefly described as follows. KTH dataset is one  of the known event recognition datasets for human actions, which introduces a video database containing six types of human actions (walking, jogging, running, boxing, hand waving, and hand clapping) performed by $25$ people. The sequences are downsampled to the spatial resolution of $160 x 120$ pixels and have a length of four seconds in average with $25$ fps frame rate~\cite{Schuldt2004}.  The Weizmann dataset is another human action dataset, it contains a database of $90$ low-resolution ($180 x 144$ pixels) with $50$ fps frame rate. The video sequences performed by 9 people, each one performing $10$ natural actions such as (run, walk, skip, jumping-jack, jump-forward-on-two-legs, jump-in-place-on-two-legs, gallopsideways (side), wave-two-hands, waveone-hand, and bend)~\cite{Blank2005}.

\section*{Acknowledgments}

This research is supported by the Center of Research Excellence in Hajj and Omrah (HajjCoRE) at Umm Al-Qura University in Makkah, KSA, under a project number P1127, entitled "Crowd-Sensing System: Crowd Estimation, Safety, and Management", 2011-2012.
\bibliographystyle{plain}

\bibliographystyle{ieeetr}

\end{document}